\documentclass{Interspeech2024}
\usepackage{multirow}
\usepackage{subcaption}

\interspeechcameraready

\title{CrisperWhisper: Accurate Timestamps on Verbatim Speech Transcriptions}

\name[affiliation={1}]{Laurin}{Wagner}
\name[affiliation={1}]{Bernhard}{Thallinger}
\name[affiliation={1}]{Mario}{Zusag}

\address{
  $^1$nyra health}
\email{lwagner@nyra.health, bthallinger@nyra.health, mzusag@nyra.health}


\begin{document}

\maketitle

\begin{abstract}
    We demonstrate that carefully adjusting the tokenizer of the Whisper speech recognition model significantly improves the precision of word-level timestamps when applying dynamic time warping to the decoder's cross-attention scores. We fine-tune the model to produce more verbatim speech transcriptions and employ several techniques to increase robustness against multiple speakers and background noise. These adjustments achieve state-of-the-art performance on benchmarks for verbatim speech transcription, word segmentation, and the timed detection of filler events, and can further mitigate transcription hallucinations. The code is available open source\footnote{\url{https://github.com/nyrahealth/CrisperWhisper}}.
\end{abstract}

\section{Introduction}
Training deep-learning models on large-scale, weakly supervised speech datasets have proven very effective for extracting rich representations, which perform well on versatile speech processing tasks, such as automatic speech recognition (ASR) \cite{wav2vec2, wavlm, hubert} or speaker verification \cite{bigssl, kang2022augmentation}. Notably, Radford et al. \cite{original_whisper} trained Whisper, a sequence-to-sequence (Seq2Seq) transformer model \cite{transformer} on 680,000 hours of weakly supervised speech recognition data, demonstrating strong generalisation abilities across domains, languages and datasets.

Recent works \cite{whisper_not_verbatim, romana2023automatic} show that Whisper eliminates many filler words, recurring utterances and other artifacts, which Lea et al. \cite{lea2023user} refer to as an \textit{intended} transcription style, suitable for contexts where clarity of intent is prioritized over detailed speech analysis. This style, however, does not support the detection, categorization, or analysis of disfluencies and therefore omits many clinically relevant aspects of speech. \textit{Verbatim} speech transcriptions capture all articulated utterances and can efficiently be used for clinical assessment of speech \cite{romana2021automatically, wagner2023careful}. A speech disfluency occurs when there's an interruption in the normal rhythm of speech, typically manifesting as filled pauses, word repetitions, or corrections. Filled pauses, beyond their linguistic interest \cite{clark2002using}, provide insight into the language planning process and are indicators of cognitive load \cite{cognitiveLoad, cl2, article_fillers}. Therefore, analyzing the timing and frequency of disfluencies, particularly common fillers \cite{womack-etal-2012-disfluencies} such as 'uh' and 'um', offers valuable insights into a speaker's cognitive processes. Many clinically relevant biomarkers like speech rate or productive time ratio \cite{wagner2023careful} rely on time accurate detection of all aspects of speech. Wagner et al. \cite{wagner2023careful} show that fluency markers derived from timing information alone are sufficient to differentiate between four different aphasia sub-types with an \(F_1\)-score of 81.6. 
Ge et al. \cite{Zhu:FillerWords:INTERSPEECH:22} developed a filled pause detection dataset and a pipeline combining ASR, a Voice Activity Detection (VAD) model, and a classifier. Speech regions that remained untranscribed by the ASR model are fed to a VAD model. The resulting voice-active regions are then further classified to identify filled pause events. This approach outperforms a convolutional recurrent neural network (CRNN), which operates on log-mel spectograms with 128 bins computed from 1 second clips and a forced-aligner based method  called Gentle \cite{gentle} combined with an acoustic model. However, distinguishing between filler words and other disfluency types such as false starts, remains challenging for this uncontextualized system. In contrast, Whisper's contextual capabilities could offer improved speech analysis capabilities in noisy scenarios.

Whisper does not provide word-level timestamps natively. To this end, WhisperX \cite{bain2023whisperx} uses force-alignment between Whisper's transcriptions and a connectionist temporal classification (CTC) based phoneme model. This forced phoneme alignment transfers the timing information from the CTC-based Wav2Vec2.0 model \cite{wav2vec2} onto Whisper's transcripts. Their VAD-based cut and merge approach allows for segmenting audio efficiently before transcription, improving both speed and accuracy of the transcriptions. However, this method faces challenges, since discrepancies between model transcripts can further degrade timestamp precision. Additionally, employing a second model increases complexity and the VAD-based segmentation approach, while efficient, lacks robustness in noisy environments. Moreover, Wav2Vec2.0 tends to be less noise robust than Whisper, further degrading timestamps in noisy scenarios. Another approach that gained popularity for inferring word-level timestamps uses Dynamic Time Warping(DTW) on the cross-attention scores of the Whisper decoder \cite{openai-dtw, lintoai2023whispertimestamped}. 
Further, in the context of disfluencies, Koenecke et al. \cite{koenecke2024careless} examined Whisper's issues of producing hallucinated content when transcribing speech from people with aphasia. Utilizing samples of AphasiaBank \cite{macwhinney2011aphasiabank}, they showed that roughly 1\% of the produced transcripts contained hallucinated content. 

We show that by adjusting Whisper's tokenizer and carefully fine-tuning Whisper on artificial perturbations for noise robustness and single-speaker focus (i) the word-level timestamps can be improved significantly using a single model (ii) the verbatim transcription style reaches state-of-the-art results on more verbatim datasets such as the AMI Meeting Corpus \cite{carletta2005ami} or TED-LIUM \cite{hernandez2018ted}, while maintaining the same accuracy on datasets such as Librispeech \cite{panayotov2015librispeech} or CommonVoice \cite{commonvoice:2020} (iii) the model achieves near perfect filled pause detection accuracy and (iv) we can substantially mitigate hallucinations. We call the resulting model CrisperWhisper for its \textit{crisp} timestamps and are open-sourcing a synthetic dataset with accurate word-level timestamps as well as the code for CrisperWhisper. 

\section{CrisperWhisper}
\subsection{DTW for Timestamp Prediction}
\subsubsection{Intuition}
Whisper employs an encoder-decoder structure, where the encoder incorporates multiple Transformer encoder blocks to process audio. Initially, audio is re-sampled to \SI{16}{\kilo\hertz} and converted into an $80$-channel log-magnitude Mel spectrogram and downsample via convolutions. The encoder operates on these downsampled spectograms in \SI{25}{\milli\second} windows with a stride of \SI{20}{\milli\second}, meaning that each processed state represents \SI{25}{\milli\second} of audio, which is shifted by \SI{20}{\milli\second} steps. The Whisper large model series uses a byte-level Byte Pair Encoding (BPE) text tokenizer \cite{gage1994new}, which produces the targets during training. Whisper's Transformer decoder uses cross-attention layers. The resulting cross-attention scores effectively reflect the decoder's focus on specific segments of the encoder output during the token prediction process. The intuition is that this focus is indicative of the decoder's strategy to prioritize encoder output regions most relevant to the current token's prediction. The goal is therefore to use the network's cross attention scores to assess which \SI{25}{\milli\second} audio frames were important to decode the current token and use the aggregate of these frames as a timestamp \cite{lintoai2023whispertimestamped}.

\subsubsection{DTW and Cost Matrix Construction}
We employ the DTW \cite{dtw} algorithm to find the optimal cost, monotonic and continuous alignment between two sequences, requiring a cost matrix to measure the alignment expense between elements of these sequences. In our case these sequences are the encoder outputs \(E = \{e_1, e_2, \ldots, e_n\}\) representing the encoded acoustic signal and the sequence of decoder token predictions \(D = \{d_1, d_2, \ldots, d_m\}\). Given a set of suitable attention heads from the decoder \(H = \{h_1, h_2, \ldots, h_l\}\) the cost matrix is defined as follows. Each of the  \(d_i\) is associated with a set of cross-attention vectors \(A_i = \{a_{i1}, a_{i2}, \ldots, a_{il}\}\), where \(a_{ik}\) denotes the attention score from the \(k\)-th attention head when decoding the \(i\)-th token, with each \(a_{ik} \in \mathbb{R}^n\). We average these vectors (\(\bar{A}_i = \frac{1}{l} \sum_{k=1}^{l} a_{ik}\)) and normalize them to construct the cost matrix \(C\):
\[
C := - \begin{bmatrix}
\frac{\bar{A}_1}{\left\lVert \bar{A}_1 \right\rVert_2} \\
\vdots \\
\frac{\bar{A}_m}{\left\lVert \bar{A}_m \right\rVert_2}
\end{bmatrix}
\]
Crucially and in contrast to \cite{lintoai2023whispertimestamped}, we remove all tokens corresponding to punctuation from \(D\) before constructing the cost matrix since punctuation has no clear acoustic representation and should therefore not be given a timestamp in the alignment. 

\subsection{Retokenization}
\begin{figure}[ht]
  \centering
  \includegraphics[width=\linewidth]{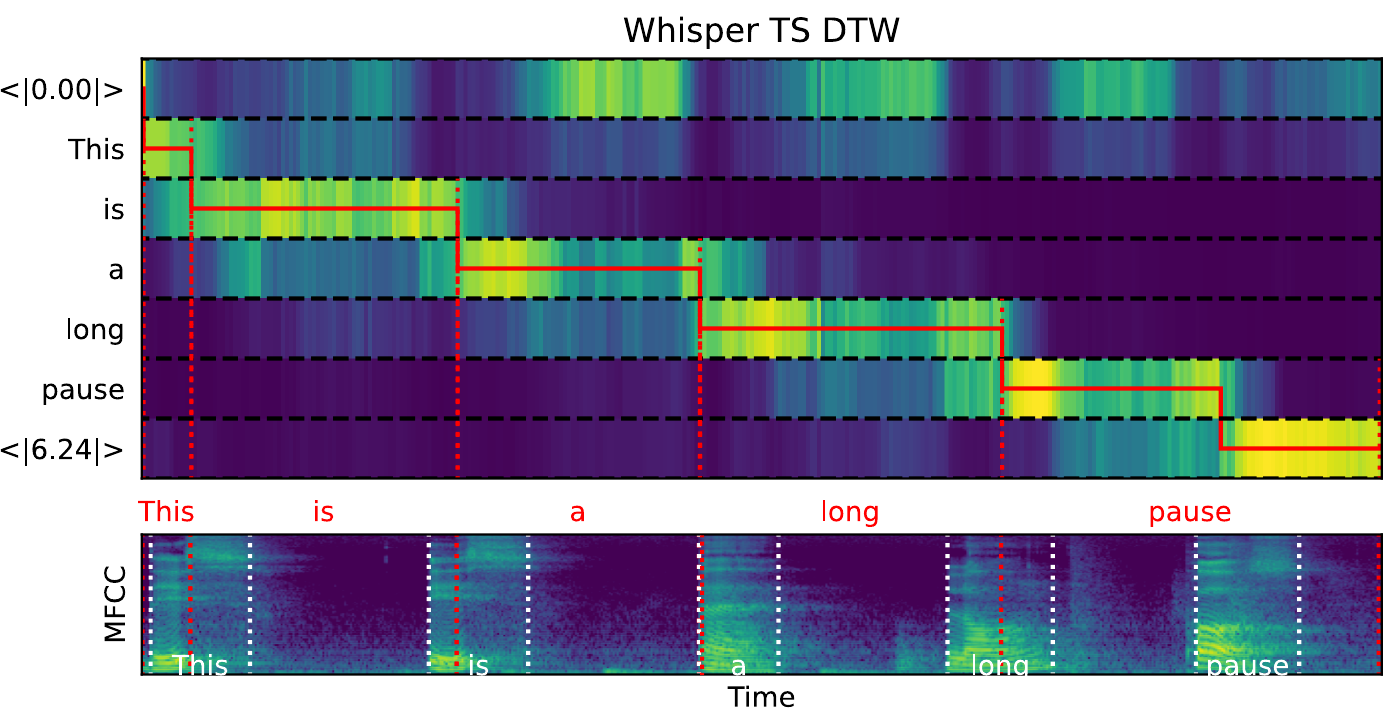}
  \caption{Example of a DTW path through the cross-attention weights matrix of Whisper large-v2 as in \cite{lintoai2023whispertimestamped}. White lines represent the ground truth.}
  \label{fig:wt_dtw}
\end{figure}

\begin{figure}[ht]
  \centering
  \includegraphics[width=\linewidth]{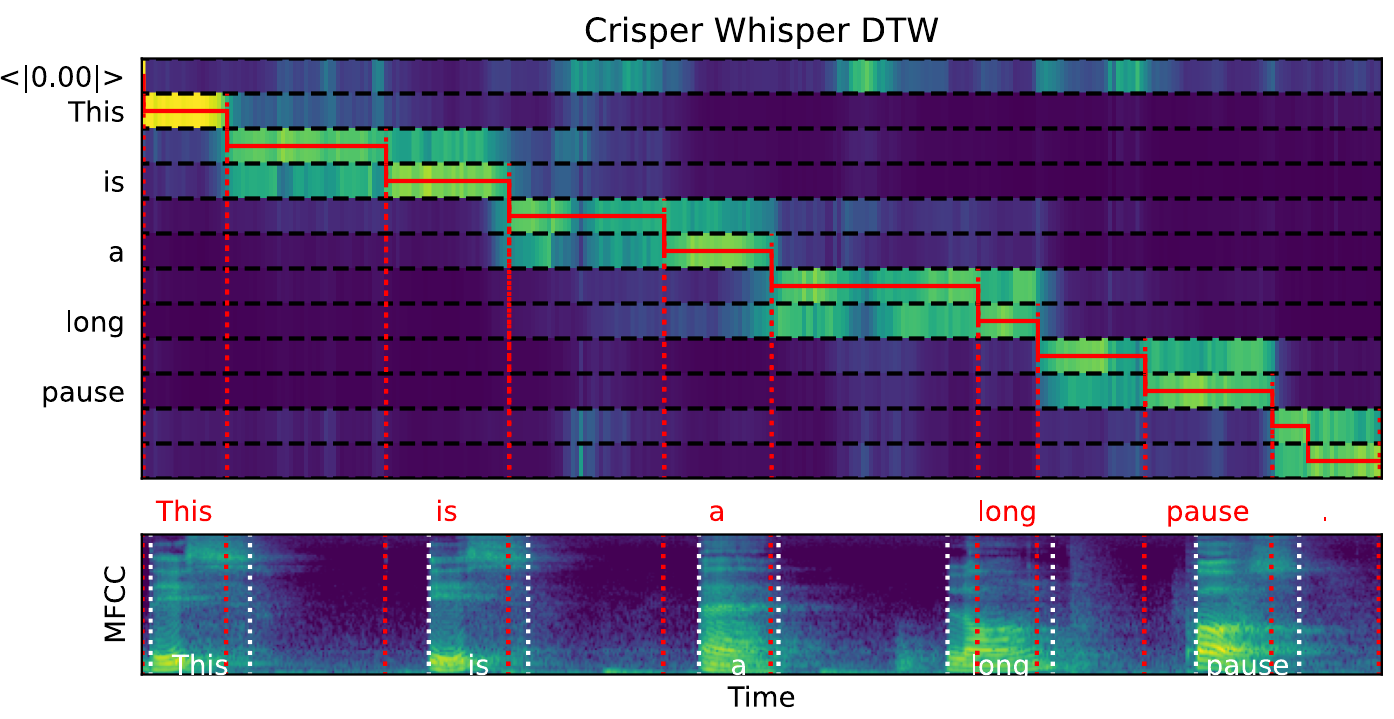}
  \caption{Example of a DTW path through the cross-attention weights matrix after CrisperWhisper retokenization. White lines represent the ground truth.}
  \label{fig:cw_dtw}
\end{figure}

Taking a closer look at the tokens in the vocabulary of Whisper, we identify that many tokens are prefixed with a space. When applying the BPE algorithm to  all CommonVoice14 transcripts \cite{commonvoice:2020} using Whisper's vocabulary, we observe that only $13$\% of spaces in the original transcripts are mapped to the explicit space token. For instance, tokenizing the sentence 'This is a long pause.' with Whisper's original tokenizer results in ['This', ' is', ' a', ' long', ' pause', '.'], where spaces are included at the start of tokens rather than as standalone entities. This tokenization approach impacts the application of DTW for aligning audio segments to tokens, as it inadvertently integrates pauses at the beginning of tokens into their timings. We observe that spaces are exclusively found at the beginning of tokens but never at the end or in the middle. Therefore, we propose to simply strip all tokens in the vocabulary of spaces, except the space token itself, and keep only the unique tokens. We adjust the merges in the tokenizer to be congruent with this reduced vocabulary. This simple adjustment ensures that all spaces will be tokenized individually, theoretically enabling the DTW algorithm to detect pauses between words. Retokenizing our example with the adjustment yields ['This', ' ', 'is', ' ', 'a', ' ', 'long', ' ', 'pause', '.']. The difference between the DTW paths on the cross attention scores of Whisper's large-v2 version with its original tokenizer can be found in Figure \ref{fig:wt_dtw} with the fine-tuned CrisperWhisper model and its adjusted tokenizer in Figure \ref{fig:cw_dtw}, visualising the close alignment with the ground truth timings. We further re-purpose tokens for 'uh' and 'um' to canonically transcribe filled pause events.

\subsection{Pause Heuristics}
To address the overestimation of pause durations by the DTW algorithm due to non-distinct attentions, we introduced a heuristic that splits the duration of pauses evenly between the preceding and subsequent words, setting a cap at \SI{160}{\milli \second}. This cap is based on the observed distribution of pause durations, effectively distinguishing between insubstantial "artifact" pauses and meaningful speech pauses. Durations surpassing this cap are identified and timed as genuine pauses, ensuring a more accurate representation of speech rhythm.

\section{Training}
\subsection{Datasets}
\label{sec:train_ds}

Our training dataset consists of two spontaneous speech datasets with a verbatim transcription style, namely the \textbf{AMI Meeting Corpus} \cite{carletta2005ami} and a specially adapted version of the \textbf{PodcastFillers Corpus} \cite{Zhu:FillerWords:INTERSPEECH:22}, along with a cleaned segment of the \textbf{CommonVoice14 Corpus} \cite{commonvoice:2020} English subset. Additionally, we use two noise datasets, \textbf{FSDnoisy18k} \cite{freesound_noise} and \textbf{AudioSet} \cite{audioset}, to make the model more noise invariant. Specifically, for the AMI dataset, we utilize the training split of the AMI-IHM subset, which contains approximately 29,000 meeting recording clips with canonical transcriptions of filler events. Moreover, we utilize a subset of the PodcastFillers dataset, which comprises approximately 35,000 instances of filler words such as 'uh' or 'um' used in podcast episodes. The dataset includes timings and automatically generated timed transcriptions for the podcast episodes. Our process for reformatting this dataset involves the following steps:

\begin{enumerate}
    \item \textbf{Sampling Context:} For each timed filler word in the training set, we generate three distinct audio segments by choosing varying context lengths ranging from 1 to 5 seconds from both before and after the filler. This selection is made with care to avoid including partial words on both ends, slightly adjusting the sampled context length when needed to include partial words fully. This expands our dataset to approximately 105,000 samples.
    
    \item \textbf{Cut Audio Clips with Aligned Transcripts:} The chosen audio segments, along with their aligned transcripts, are extracted. Filler words are explicitly marked as either 'uh' or 'um' in their respective positions within the transcripts.
    
    \item \textbf{Transcript Correction:} To address inaccuracies in punctuation and capitalization within the original transcripts, we utilized GPT-4 \cite{achiam2023gpt} after observing that the original transcript quality adversely affected Whisper's ability to correctly apply punctuation.
    
\end{enumerate}

For CommonVoice14, we employed Whisper's medium model on the English train subset to identify and remove samples likely not transcribed verbatim. Any sample with a character error rate exceeding  $3$\% compared to its label was excluded. 

\subsection{Implementation Details}
To deploy CrisperWhisper as a comprehensive speech analysis model in practical applications, it is essential that the model is trained to prioritize the primary speaker's voice and to be generally noise robust. To this end we use the noisy/overlapped speech simulation proposed in WavLM \cite{wavlm} during fine-tuning from the whisper-large-v2 checkpoint \cite{openai_whisper_large_v2}. As noise data, we use FSDnoisy18k, AudioSet, random Gaussian noise and random speech samples drawn from the dataset. To counteract hallucinations, we introduce noise-only samples (containing no speech) with empty transcriptions in 1\% of the training samples. The training process spans 6,000 steps with a batch size of 256, utilizing a 0.00005 learning rate, a linear learning rate decay with an 800-step warmup phase, amounting to approximately 2 epochs.

\section{Evaluation}
\label{sec:eval}

\subsection{Datasets}
\label{sec:eval_datasets}
\textbf{AMI Meeting Corpus}: We use the official test split on the AMI-IHM subset with approximately 11,500 samples. \textbf{AMI disfluency subset}: The AMI Meeting Corpus contains filler words transcribed in a canonical way. We extend the filler words by using GPT-4 to label repetitions, false starts and revisions on the transcripts of the AMI-IHM test set. Our disfluency subset consists of all files and transcripts that contain at least one of the labeled disfluencies and contain more than 5 words, which are approximately 4,000 samples. \textbf{PodcastFillers Corpus}: We use the same approach as described in Section \ref{sec:train_ds} for creating annotated filler samples on the official test subset, choosing 1 second of context before and after each filler, which results in approximately 5,000 samples. \textbf{Synthetic dataset}: To compare word-level timestamp accuracy, we created 200 samples of spontaneous speech transcripts with GPT-4, which contain natural pauses in sentences, indicated by '...' in the transcripts. These transcripts were subsequently synthesized with ElevenLabs \url{https://www.elevenlabs.io}, creating naturally sounding spontaneous speech samples for which timestamps were manually annotated. \textbf{AphasiaBank Corpus}: AphasiaBank \cite{macwhinney2011aphasiabank} comprises a collection of interviews between clinicians and subjects afflicted with aphasia, as well as healthy control subjects. We are using the same files that Koenecke et al. \cite{koenecke2024careless} have identified to cause hallucinated content when transcribed with Whisper large-v2. \textbf{TED-LIUM}: We use the test split of TED-LIUM Release 3 \cite{hernandez2018ted}, using the segmented manual test transcripts included in the release. \textbf{LibriSpeech}: We use both of the popular LibriSpeech  \cite{panayotov2015librispeech} 'test clean' and 'test other' splits for evaluation.

\subsection{Metrics}
\label{sec:metrics}
To evaluate transcription accuracy, we use word error rate (\textbf{WER}) and insertion error rate (\textbf{IER}) to quantify word omissions. For timing accuracy, we use the \textbf{F\textsubscript{1}-score}, which is well defined via basic confusion matrix terminology. In this context, we define a \textbf{true positive} as a predicted word that both overlaps temporally with a reference word and matches its content. Each reference word can only contribute to a single true positive. A \textbf{false positive} is defined as a predicted word that does not have temporal overlap or content match with any reference word. Conversely, a \textbf{false negative} is a reference word that does not have temporal overlap or content match with any predicted word. Temporal overlap occurs when the start (onset) and end (offset) timestamps of a prediction fall within a predefined collar of the corresponding timestamps in the reference. Additionally, we evaluate localization accuracy using the \textbf{mean Intersection over Union (mIoU)} metric, which compares each predicted word against the reference for both string match and temporal overlap, calculating the IoU based on timestamps, or assigning a score of 0 if no match exists. The highest IoU score for each word represents its IoU, ensuring each word is matched only once.

\subsection{Results}
In the following, we are referring to Whisper's large-v2 model with the DTW implementation of \cite{lintoai2023whispertimestamped} as \textbf{WhisperT}, the default configuration of WhisperX \cite{bain2023whisperx} with an underlying Whisper large-v2 and Wav2vec2.0 alignment model as \textbf{WhisperX}.

\subsubsection{Word Segmentation Performance}
\begin{figure}[ht!]
    \centering
    \begin{subfigure}[b]{0.49\linewidth}
        \centering
        \includegraphics[width=\linewidth]{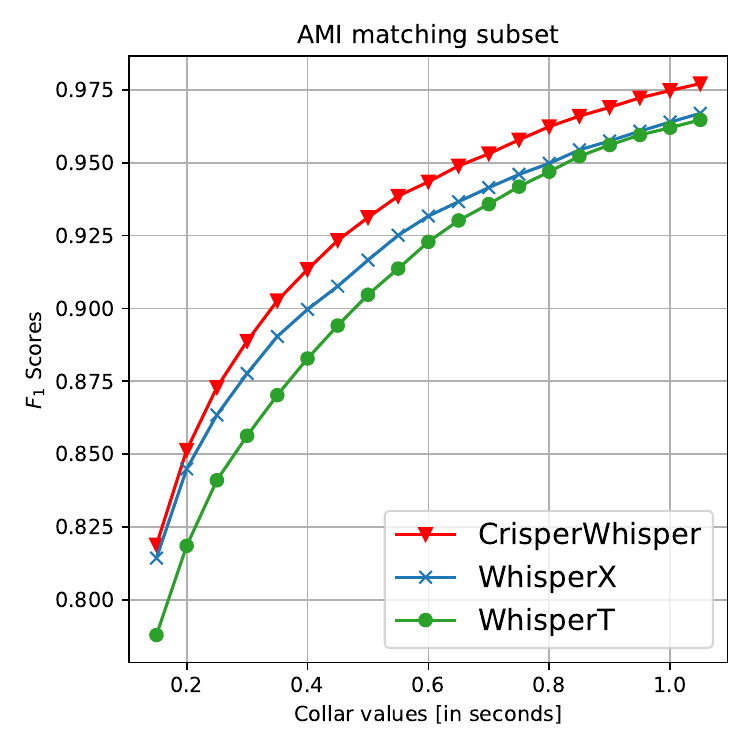}
        \caption{Matching AMI subset}
        \label{fig:ami_matching_subset}
    \end{subfigure}
    \hfill 
    \begin{subfigure}[b]{0.49\linewidth}
        \centering
        \includegraphics[width=\linewidth]{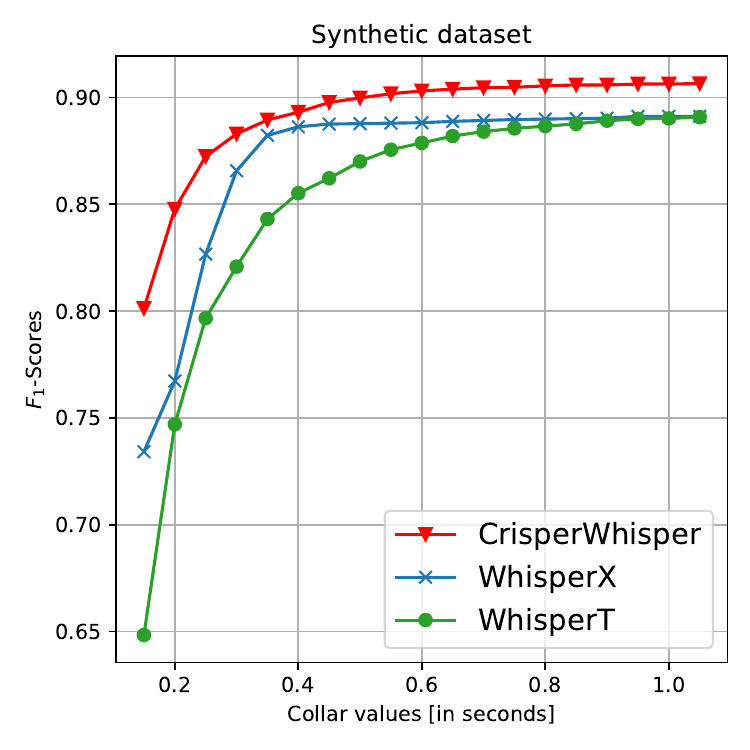}
        \caption{Synthetic dataset}
        \label{fig:synthetic_dataset}
    \end{subfigure}
    \caption{Word segmentation performance showing the F\textsubscript{1}-score for different collar values.}
    \label{fig:test}
\end{figure}

Figure \ref{fig:ami_matching_subset} shows how CrisperWhisper outperforms previous state-of-the-art models using different collar values on the test set of the AMI Corpus. We ensure that the normalized transcripts of our prediction, the normalized reference and the normalized predictions of the other models coincide completely. This is to account for the fact that our more verbatim approach gives us an unfair advantage and we want to ensure that we evaluate the localization performance separately from the transcription accuracy. Figure \ref{fig:synthetic_dataset} depicts the segmentation performance using different collar values on the clean, synthetic dataset mentioned in Section \ref{sec:eval_datasets} with manually annotated timestamps. 

We further evaluate the noise robustness of our model by adding random secondary voice samples from the LibriSpeech 'test clean' subset, white noise, and random noise samples from FSDnoisy18k with a signal-to-noise ratio of 1:5 to the synthetic samples. As detailed in Table \ref{table:synthetic_data}, CrisperWhisper demonstrates superior robustness in terms of mIoU and F\textsubscript{1}-score under noisy conditions. In contrast, WhisperX exhibits a more significant performance decline than WhisperT, attributable to Wav2Vec2.0's lesser noise resilience. Notably, CrisperWhisper achieves markedly higher mIoU metrics and F\textsubscript{1}-scores, particularly with narrower collars, underscoring its enhanced accuracy in timestamping pauses compared to other evaluated methods.

\begin{table}[th!]
   
    \caption{Noise robustness of word segmentation performance on synthetic data using a collar of 0.2 seconds.}
    \label{table:synthetic_data}

    \centering
    \begin{tabular}{lcccccc}
        \toprule
        \textbf{Model} 
        & \multicolumn{2}{c}{\textbf{Synthetic}} 
        & \multicolumn{2}{c}{\textbf{Synthetic noisy}} \\
        \cmidrule(lr){2-3}
        \cmidrule(lr){4-5}
        & $F_1$ $\uparrow$ & mIoU $\uparrow$ 
        & $F_1$ $\uparrow$ & mIoU $\uparrow$ \\
        
        \midrule
        WhisperT \cite{openai_whisper_large_v2} 
        & 74.7 & 51.4 
        & 68.3 & 49.8  \\
        WhisperX \cite{bain2023whisperx} 
        & 76.7 & 61.5 
        & 59.0 & 44.3 \\
        \midrule
        CrisperWhisper 
        & \textbf{84.7} & \textbf{63.4} 
        & \textbf{79.5} & \textbf{60.5} \\
        
        \bottomrule
    \end{tabular}

\end{table}

\subsubsection{Disfluency Segmentation Performance}
\begin{figure}[ht!]
    \centering
    \begin{subfigure}[b]{0.49\linewidth}
        \centering
        \includegraphics[width=\linewidth]{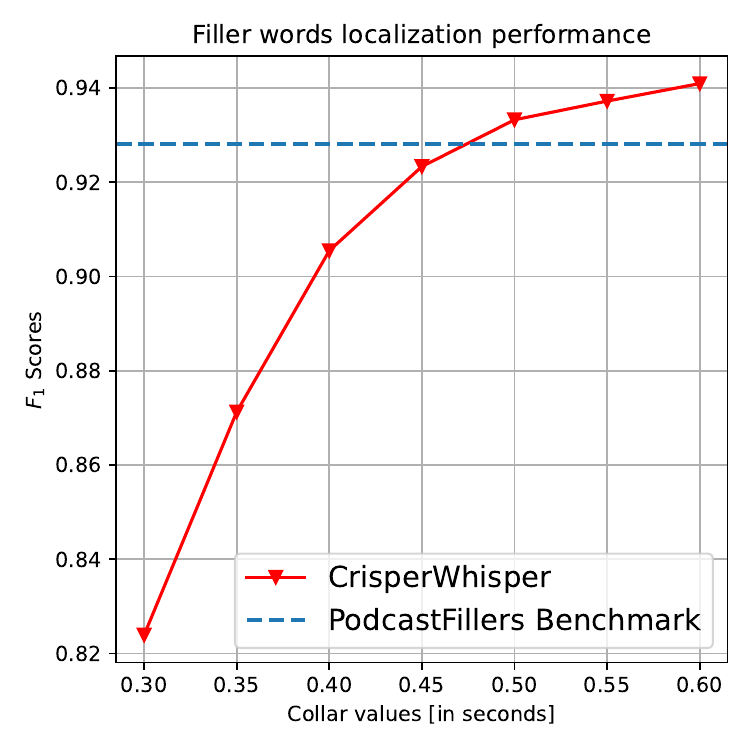}
        \caption{PodcastFillers Dataset}
        \label{fig:filler_localization}
    \end{subfigure}
    \hfill 
    \begin{subfigure}[b]{0.49\linewidth}
        \centering
        \includegraphics[width=\linewidth]{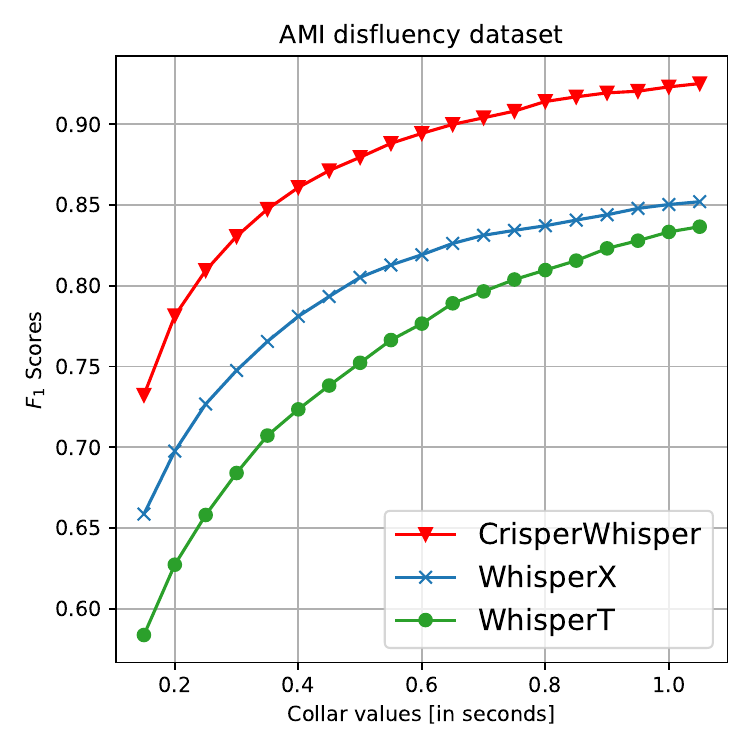}
        \caption{AMI disfluent subset}
        \label{fig:disfluency_subset}
    \end{subfigure}
    \caption{Disfluency Localization Performance}
    \label{fig:disfluency_detection}
\end{figure}

For evaluating the filler word detection and segmentation performance, we transcribe the audio examples of our adjusted test split of the Podcast-Fillers Corpus as described in Section \ref{sec:eval} with CrisperWhisper and calculated the $F_1$-scores as described in Section \ref{sec:metrics} for various collar values. The results can be seen in Figure \ref{fig:filler_localization} with the reported \(F_1\)-score eventually exceeding the acoustic model reported in \cite{Zhu:FillerWords:INTERSPEECH:22}, although the segmentation is worse for collar values smaller than 0.5 seconds. Since our model transcribes verbatim, we also detect and segment other disfluency types, such as repetitions, false starts or partial words. Figure \ref{fig:disfluency_subset} shows the localization performance on disfluent speech samples of the AMI disfluency subset described in Section \ref{sec:eval_datasets}. 

\subsubsection{Verbatim Transcription Performance}
Table \ref{table:asr_results} shows the significantly improved ASR performance on spontaneous speech datasets with more verbatim transcriptions. We have further validated that CrisperWhisper's transcription accuracy does not degrade on intended speech datasets. All transcriptions of the 'test other' and 'test clean' subsets of the LibriSpeech Corpus or the test split of CommonVoice14 lie within 0.01 WER of the original Whisper large-v2 model.

\begin{table}[th!]
   
    \caption{ASR performance on verbatim datasets.}
    \label{table:asr_results}

    \centering
    \begin{tabular}{lcccccc}
        \toprule
        \textbf{Model} 
        & \multicolumn{2}{c}{\textbf{AMI}} 
        & \multicolumn{2}{c}{\textbf{TED-LIUM}} \\
        \cmidrule(lr){2-3}
        \cmidrule(lr){4-5}
        & WER $\downarrow$ & IER $\downarrow$ 
        & WER $\downarrow$ & IER $\downarrow$ \\
        
        \midrule
        Whisper \cite{openai_whisper_large_v2} 
        & 16.82 & 11.77 
        & 4.01 & 3.08  \\
        \midrule
        CrisperWhisper 
        & \textbf{9.72} & \textbf{2.26} 
        & \textbf{3.26} & \textbf{0.75} \\

        \bottomrule
    \end{tabular}

\end{table}

\subsubsection{Hallucinations}
To validate hallucination mitigation, we use the same audio files from AphasiaBank as analyzed by Koenecke et al. \cite{koenecke2024careless}. CrisperWhisper does not produce harmful hallucinations on any of the identified speech recordings, but produces repetitive transcription loops on 10 recordings. Since CrisperWhisper is producing accurate word-level timestamps, we are simply removing tokens with a duration below $50$ \si{ms}, effectively eliminating this type of artefact of hallucinating speech during inactivate speech regions.

\section{Conclusion}
We proposed CrisperWhisper, a robust end-to-end speech transcription model producing accurate word-level timestamps in a verbatim, single speaker focused transcription style. We trace the problem of unsharp timestamps around disfluencies and pauses back to Whisper's tokenizer and present a strategy to alleviate this problem. One weakness of our approach is the arbitrary selection of attention heads used for alignment. In the near future, we want to investigate the verbatim transcription and segmentation capabilities for quantifying speech deficits, scale the approach with more high quality verbatim data and explore how these capabilities can be transferred to other languages.

\newpage

\bibliographystyle{IEEEtran}
\bibliography{mybib}

\end{document}